\documentclass[runningheads]{llncs}
\usepackage[T1]{fontenc}
\usepackage{graphicx}

\usepackage[whole]{bxcjkjatype} 
\usepackage{multirow}
\usepackage{booktabs} 

\usepackage{amsmath}
\usepackage{array}
\usepackage[table]{xcolor}

\usepackage{tikz}
\usepackage{tcolorbox}
\tcbuselibrary{skins}
\tcbuselibrary{breakable}

\newtcolorbox{simplebox}{
  enhanced,
  drop fuzzy shadow,
  colback=white,
  boxrule=0.5pt
}

\newtcolorbox{expbox}{
  enhanced,
  colback=white,
  boxrule=0.5pt,
  breakable
}

\newtcolorbox{alertbox}{
  enhanced,
  drop lifted shadow,
  sharp corners,
  colback=white,
  colbacktitle=white,
  coltitle=black,
  boxrule=1.0pt,
  breakable
}

\newtcolorbox{alertboxa}[1]{
  title={#1},
  enhanced,
  drop lifted shadow,
  sharp corners,
  colback=white,
  colbacktitle=black!10!white,
  coltitle=black,
  boxrule=1.0pt,
  breakable
}

\newtcolorbox{quoteboxa}[1]{
  title=#1,
  enhanced,
  colback=white,
  colbacktitle=white,
  coltitle=black,
  boxrule=0.8pt,
  attach boxed title to top left={xshift=3mm, yshift*=-\tcboxedtitleheight/2},
  box align=top
}

\newcommand{\syllogism}[3]{
\begin{tabular}{l}
\textsf{Major premise}: #1\\
\textsf{Minor premise}: #2\\ \hline
\textsf{Conclusion}: #3
\end{tabular}
}

\newcommand{\abduction}[3]{
\begin{tabular}{l}
\textsf{Major premise}: #1\\
\textsf{Conclusion}: #2\\ \hline
\textsf{Minor premise}: #3
\end{tabular}
}

\newcommand{\abductionCall}[3]{
\begin{tabular}{l}
\textsf{Rule}: #1\\
\textsf{Observation}: #2\\ \hline
\textsf{Hypothesis}: #3
\end{tabular}
}

\newcommand{\abductionC}[3]{
\begin{tabular}{l}
\textsf{R}: #1\\
\textsf{O}: #2\\ \hline
\textsf{H}: #3
\end{tabular}
}

\newcommand{\deductionC}[3]{
\begin{tabular}{l}
\textsf{P1}: #1\\
\textsf{P2}: #2\\ \hline
\textsf{C}: #3
\end{tabular}
}

\newcommand{\corC}{yellow!30}
\newcommand{\incorC}{gray!30}

\begin{document}
\title{Abductive Reasoning with Syllogistic Forms \\
in Large Language Models}

\titlerunning{Abductive Reasoning in Large Language Models}

\author{
Hirohiko Abe\inst{1} \and
Risako Ando\inst{1} \and
Takanobu Morishita\inst{1} \\
Kentaro Ozeki\inst{1,2} \and
Koji Mineshima\inst{1} \and
Mitsuhiro Okada\inst{1}
}
\authorrunning{H. Abe et al.}
\institute{Keio University, Tokyo, Japan \and
The University of Tokyo, Tokyo, Japan
\email{hirohiko-abe@keio.jp,risakochaan@keio.jp,kentaro.ozeki@gmail.com}\\
\email{morishita@keio.jp,\{minesima,mitsu\}@abelard.flet.keio.ac.jp}
}

\maketitle

\begin{abstract}
Research in AI using Large-Language Models (LLMs) is rapidly evolving, and the comparison of their performance with human reasoning has become a key concern. Prior studies have indicated that LLMs and humans share similar biases, such as dismissing logically valid inferences that contradict common beliefs. However, criticizing LLMs for these biases might be unfair, considering our reasoning not only involves formal deduction but also abduction, which draws tentative conclusions from limited information. Abduction can be regarded as the inverse form of syllogism in its basic structure, that is, a process of drawing a minor premise from a major premise and conclusion.
This paper explores the accuracy of LLMs in abductive reasoning by converting a syllogistic dataset into one suitable for abduction. It aims to investigate whether the state-of-the-art LLMs exhibit biases in abduction and to identify potential areas for improvement, emphasizing the importance of contextualized reasoning beyond formal deduction. This investigation is vital for advancing the understanding and application of LLMs in complex reasoning tasks, offering insights into bridging the gap between machine and human cognition.

\keywords{Abduction  \and Deduction \and Syllogism \and Reasoning bias \and Large Language Models.}
\end{abstract}

\section{Introduction}
Research in Large-Language Models (LLMs) is rapidly advancing, with a significant focus on comparing their performance to human reasoning. 
Previous studies have shown that while LLMs generally excel at reasoning tasks~\cite{brown2020language,wei2022chain,kojima2022large}, they exhibit similar biases to humans, such as dismissing logically valid inferences that contradict common beliefs~\cite{dasgupta2023language,ando-etal-2023-evaluating}. 
Although these studies often emphasize deductive reasoning, our everyday reasoning encompasses more than just deduction. 
Given that LLMs are developed by learning natural language used in daily contexts without specialized logical training, it would be unreasonable to criticize them for bias tendencies in deduction tasks.
Since our reasoning involves not only formal deduction but also abduction, which draws hypotheses from limited information, it is crucial to investigate LLMs' capabilities in making abductive inferences.

Abduction is a natural form of reasoning that seeks reasons and explanation.
For example, when discussing the reason for the delay of the train by asking ``Why was the train late?'' it is natural to trace back from the observed fact to the reason explaining it, such as ``because the traffic lights failed.'' However, it is rare to ask in the form of a deduction, as in ``The traffic lights failed, therefore the train was late.''
The explanatory aspect of abduction, as well as the logical consistency in deductive reasoning, is important in investigating natural explanations and realizing an explainable AI (XAI) that naturally answers \textit{why}-questions~\cite{Medianovskyi2022-MEDOEA-2}.
Investigating how accurately current LLMs can perform abduction provides a theoretical basis for research into XAI.

Abduction is important in knowledge acquisition.
Charles Sanders Peirce~\cite{peirceCollectedPapers} regarded abduction as a process of inquiry along with deduction and induction.
Abduction plays a more important role than deduction, especially when it comes to the discovery of the unknown.
Evaluating LLMs' abductive reasoning abilities is essential in determining whether LLMs can gain new knowledge, particularly from limited information.

Inquiry, or the activity of acquiring knowledge, is considered one of the chief topics in recent epistemology, and norms of inquiry have been studied~\cite{Hookway2006-HOOEAI,Hookway2008-HOOQEA,FriedmanForthcoming-FRIZE}
Given that abduction is related to inquiry in Peirce's~\cite{peirceCollectedPapers} philosophy, it can be considered a form of logic that guides inquiry, for example, by providing hypotheses and guidance on what to investigate. Assessing LLMs' abductive reasoning ability is important when addressing the question of whether LLMs can be used to guide our everyday inquiry.

In this paper, we introduce a dataset to test the abductive reasoning abilities of LLMs, compare LLMs' accuracy on abductive reasoning tasks with deductive reasoning tasks, and explore whether LLMs show human-like belief biases on reasoning.\footnote{The dataset is available at \url{https://github.com/kmineshima/abduction-syllogism-llm}.}
By LLMs, we focus on in-context learning pretrained models such as GPT~\cite{ouyang2022gpt3,openai2023gpt4} and Llama~\cite{llama3modelcard}, rather than those requiring fine-tuning such as BERT~\cite{devlin-etal-2019-bert}. These in-context learning models adapt to a specific task using a task description or a few examples of correct answers as input, called a \textit{prompt}, without changing the models' parameters.
We revealed that LLMs generally performed more poorly on abductive tasks compared to deductive tasks. In addition, we found that LLMs exhibit human-like belief biases in both abduction and deduction.

With regard to the comparison between deduction and abduction, it has been pointed out that in diagnostic inference, the inference from effect to cause, there is reason to believe that the deductive model is a more natural reasoning scheme for humans than the abductive model, under a probabilistic setting~\cite{Stilgenbauer2017ReasoningSF,Stilgenbauer2019AssessingTA}.
Although this paper shares interests with these trends in the comparison between deduction and abduction, our dataset 
specifically focuses on abductions that can be generated within the framework of syllogisms, especially by swapping premises and conclusions of deductive syllogisms.
This approach enables a systematic comparison of abduction and deduction, thereby providing a foundation for exploring more complex forms of abductive reasoning, including causal and practical variations.

\section{Background}
\label{sec:background}

\subsection{Abductive Reasoning}
\label{ssec:abductive}

Our everyday reasoning involves not only deduction but also abduction, that is, the type of reasoning that hypothetically derives new information from limited information. 
Abduction is believed to be ubiquitous in ordinary our life. 
For example, abductive reasoning is considered to be operative in cognitive process and testimonial trust.
In addition, abduction is regarded as a cornerstone of scientific methodology~\cite{douven-sep-abduction,Douven2022-DOUTAO-5}.

In general, abduction is a form of reasoning that leads to a hypothesis explaining an observed fact. Abduction is considered to be of two types: hypothesis selection and hypothesis generation. The former refers to the selection of the best explanation from among several hypotheses. It is also known as \textit{Inference to the Best Explanation} (IBE). On the other hand, the latter kind involves generating a hypothesis that explains the observed fact from given observations.

Charles Sanders Peirce first introduced abduction,
distinguishing it from deduction and induction.
Peirce~\cite{peirceCollectedPapers} understood it as ``the process of forming an explanatory hypothesis'' and ``the only logical operation which introduces any new idea'' (CP 5.171).
According to Peirce, abduction is amplicative in that it adds new information besides the premises, while deduction is not.

Peirce initially organized abduction in a syllogistic framework~\cite{Bellucci2022}.
According to Peirce, abduction is made by changing the minor premise and the conclusion in a deductively valid syllogism.
The following is an example of a deductively valid syllogism, which is called a first figure syllogism.
\begin{center}
\syllogism
{All A are B}
{C is A}
{C is B}
\end{center}
By changing the minor premise and the conclusion, we obtain the following form of abduction.
\begin{center}
\abduction
{All A are B}
{C is B}
{C is A}
\end{center}
Note that this form of inference is the so-called \textit{Affirming the Consequent}, a typical instance of deductively invalid inferences.
As a formal characterization of abduction, Peirce~\cite{peirceCollectedPapers} says, ``The surprising fact, $C$, is observed. But if $A$ were true, $C$ would be a matter of course. Hence, there is reason to suspect that $A$ is true'' (CP 5.189).
In this paper, we call the first premise of abduction \textsf{Rule}, the second premise \textsf{Observation}, and the conclusion \textsf{Hypothesis}.
The following is a concrete example of abduction based on this terminology.
\begin{center}
\abductionCall
{All things that were in the bag are white.}
{These balls are white.}
{These balls were in the bag.}
\end{center}

In the context of AI and Natural Language Processing, there have been studies on evaluating machine learning (deep learning) models using abductive reasoning.
Among others, Bhagavatula et al.~\cite{bhagavatula2020abductive} focuses on the form of abduction that selects the most plausible hypothesis that explains given observations.
This form of abduction can be subsumed under the IBE type abduction as described above.
In this paper, based on Peirce's account, we instead focus on the form of abduction that is converted from syllogism, namely, one that derives a minor premise from the major premise and conclusion of a deductively valid syllogism.

\subsection{Deductive Reasoning Abilities of LLMs}
\label{ssec:deductive}

With the rapid progress in research on LLMs, the importance of assessing their reasoning ability has increased and the abilities have been researched using a variety of tasks~\cite{wang2019superglue}.
Among these, more research that compared LLMs with humans has been conducted on deductive syllogistic reasoning, which has been studied in cognitive psychology~\cite{manktelow1999reasoningEng,chater1999probability,Geurts2003-GEURWQ}.
Datasets of syllogistic reasoning tasks for LLMs are recently introduced by Dong et al.~\cite{dong2020learning}, Guebelmann et al.~\cite{gubelmann2022philosophically}, Aghahadi et al.~\cite{aghahadi2022avicenna},
and Ando et al.~\cite{ando-etal-2023-evaluating}.
However, they only focus on deductions and do not deal with other kinds of inferences including abduction.

Recent research on LLM's reasoning abilities with syllogisms showed that while LLMs generally perform well on syllogisms, 
they tend to exhibit some human-like biases~\cite{evans1989bias,pohl2012cognitive}.
More specifically, Dasgupta et al.~\cite{dasgupta2023language}
found that LLMs reason more accurately about believable or realistic situations in reasoning tasks including syllogism and Wason's selection task.
In addition, they revealed that LLMs tend to judge inferences with believable content as valid and those with the sentences that clash our commonsense belief as invalid regardless of forms of inferences, thus failing to separate \textit{forms} from \textit{contents} (the content effects).
Ando et al.~\cite{ando-etal-2023-evaluating} 
introduced a syllogism dataset called NeuBAROCO, where syllogisms are presented in both English and Japanese. They showed that LLMs exhibit reasoning biases known in the psychological studies of syllogisms, including belief biases, conversion errors, and atmosphere effects.
Ozeki et al.~\cite{ozeki2024exploring} extended the NeuBAROCO dataset  and conducted a more detailed evaluation of a wide range of models
by implementing various reasoning tasks, including those that require translating syllogisms into logical formulas and explaining the reasoning steps.

Based on these previous findings, this paper compares the reasoning abilities of LLMs in deduction and abduction and explores whether LLMs show human-like belief biases.

\begin{table}
\centering
\caption{Eight patterns of abduction. \textsf{R}: \textsf{Rule},
\textsf{O}: \textsf{Observation}, \textsf{H}: \textsf{Hypothesis}.
Those in \colorbox{\corC}{yellow} are correct abductions, while those in \colorbox{\incorC}{grey} are incorrect.
Correct abductions are those in which \textsf{H} explains why \textsf{O} is the case, given the rule \textsf{R}.}

\scalebox{0.93}{
\begin{tabular}{cccc}
\cellcolor{\corC}\abductionC{All C are B}{These A are B}{These A are C} &
\cellcolor{\incorC}\abductionC{All B are C}{These A are B}{None} &
\cellcolor{\incorC}\abductionC{All C are B}{These A are not B}{None} &
\cellcolor{\corC}\abductionC{All B are C}{These A are not B}{These A are not C}
\medskip \\ 
\cellcolor{\incorC}\abductionC{No C are B}{These A are B}{None} &
\cellcolor{\incorC}\abductionC{No B are C}{These A are B}{None} &
\cellcolor{\corC}\abductionC{No C are B}{These A are not B}{These A are C} &
\cellcolor{\corC}\abductionC{No B are C}{These A are not B}{These A are C} \\
\end{tabular}
}
\label{tab:abduction-scheme}
\end{table}

\begin{table*}[t]
\centering
\caption{
Examples abductive syllogisms labeled as 
\textit{Consistent}, \textit{Inconsistent}, and \textit{Neutral}.
The numbers in brackets show the number of each type.
}
\begin{tabular}{ll}
\toprule
\textbf{Type} & \textbf{Example} \\ \midrule
\multirow{3}{*}{\textbf{Consistent} (66)}
& \textsf{Rule}: All people that had a fun time are smiling.\\
& \textsf{Observation}: These people are smiling.\\
& \textsf{Hypothesis}: These people had a fun time.
\\ \midrule
\multirow{3}{*}{\textbf{Inconsistent} (66)}
& \textsf{Rule}: All things that are made in the sweet restaurant are spicy.\\
& \textsf{Observation}: These cakes are made in the sweet restaurant.\\
& \textsf{Hypothesis}: These cakes are spicy.
\\ \midrule
\multirow{3}{*}{\textbf{Neutral} (84)}
& \textsf{Rule}: All things that were in the bag are white.\\
& \textsf{Observation}: These balls are white.\\
& \textsf{Hypothesis}: These balls were in the bag.
\\
\bottomrule
\end{tabular}
\label{tab:type}
\end{table*}

\section{Datasets}
\label{sec:dataset}
We formulated abduction as an inference from \textsf{Rule} and \textsf{Observation} to \textsf{Hypothesis} as shown in Section~\ref{ssec:abductive}.
\textsf{Rule} consists of a sentence of the form \textit{All A are B} (Universal Affirmative) or \textit{No A are B} (Universal Negative), while \textsf{Observation} and \textsf{Hypothesis} consist of a sentence of the form \textit{These A are B} (Particular Affirmative) or \textit{These A are not B} (Particular Negative).
We classified four patterns of abductive inference, which are shown in Table~\ref{tab:abduction-scheme}.

One of the essential features of abduction is that it is amplicative.
Abduction contributes to the acquisition of new knowledge by drawing the conclusions whose content is beyond those contained in the premises.
In this respect, abduction differs from deduction. 
The patterns of abduction identified above fulfil this characteristic.
On the other hand, the greyed-out patterns in Table~\ref{tab:abduction-scheme} do not, and they are deductively valid.
Restricting the attention to syllogistic forms helps us to define what counts as correct patterns of deductive and abductive reasoning.

To construct a set of abductive inferences having these patterns,
we first created a triple $(A, B, C)$ of terms,
where $A$ is a subject term, $B$ is an observational predicate,
and $C$ is a non-observational predicate.
Observational predicates are those that can be verified through direct observation,
while non-observational predicates are those that cannot.
For example, \textit{are white} is an observational predicate, while \textit{were in the bag} is a non-observational predicate.
In this study, we manually created 27 triples like (\textit{balls, are white, were in the bag}).

By instantiating the inference patterns shown in Table~\ref{tab:abduction-scheme} with these terms, we obtained 216 problems of abductive inference in total, with 108 correct patterns and 108 incorrect patterns.
To annotate information about belief biases,
we classified each problem by three labels, \textit{consistent}, \textit{inconsistent}, or \textit{neutral}.
The problem is \textit{consistent} if \textsf{Rule} is considered to be true as inferred from our common-sense beliefs;
it is \textit{inconsistent} if \textsf{Rule} contradicts our common-sense beliefs;
if neither holds, it is \textit{neutral}.
Table~\ref{tab:type} shows some examples of each type.

In the same way, we constructed 216 patterns of corresponding deductive inferences, with 108 valid patterns and 108 invalid patterns.
We obtained instances of deduction by changing
\textsf{Observation} and \textsf{Hypothesis} in abduction.
For example, from the first example of abduction in Table~\ref{tab:type},
we obtained an instance of valid deduction,
\textit{All people that had a fun time are smiling.
These people are smiling. Therefore, These people had a fun time.}
Table~\ref{tab:deduction-scheme} shows all eight patterns of deduction in comparison with the corresponding abductions in Table~\ref{tab:abduction-scheme}.

\begin{table}
\centering
\caption{Eight patterns of deduction. \textsf{P1}: \textsf{Major Premise},
\textsf{P2}: \textsf{Minor Premise}, \textsf{C}: \textsf{Conclusion}.
Those in \colorbox{\corC}{yellow} are valid deductions, while those in \colorbox{\incorC}{grey} are invalid.}

\scalebox{0.83}{
\begin{tabular}{cccc}
\cellcolor{\incorC}\deductionC{All C are B}{These A are B}{None} &
\cellcolor{\corC}\deductionC{All B are C}{These A are B}{These A are C} &
\cellcolor{\corC}\deductionC{All C are B}{These A are not B}{These A are not C} &
\cellcolor{\incorC}\deductionC{All B are C}{These A are not B}{None}
\medskip \\ 
\cellcolor{\corC}\deductionC{No C are B}{These A are B}{These A are not C} &
\cellcolor{\corC}\deductionC{No B are C}{These A are B}{These A are not C} &
\cellcolor{\incorC}\deductionC{No C are B}{These A are not B}{None} &
\cellcolor{\incorC}\deductionC{No B are C}{These A are not B}{None} \\
\end{tabular}
}
\label{tab:deduction-scheme}
\end{table}

\section{Experiments}
\label{sec:experiment}

\subsection{Experimental Settings and Evaluated Models}
\label{ssec:models}

We conducted experiments on two tasks: the Abduction task and the Deduction task, using the dataset created by the method described in Section~\ref{sec:dataset}. 
All experiments were conducted in English. 
In each iteration, a single problem was provided as input along with a prompt, and the model's output was collected as an answer.
The performance of the LLMs was evaluated using overall accuracy and accuracy for each problem type, providing a basic assessment of their capabilities.

In our experiments, we evaluated four state-of-the-art models with varying parameter sizes: GPT-3.5~\cite{ouyang2022gpt3}, GPT-4~\cite{openai2023gpt4}, Llama-3-8B (8 billion parameters), and Llama-3-70B (70 billion parameters) \cite{llama3modelcard}.
The GPT models are closed-source, and their specific details, including the exact number of parameters, are not publicly disclosed.\footnote{The versions of the GPT models used are \texttt{gpt-3.5-turbo-0125} and \texttt{gpt-4-0613}, accessed via OpenAI's API.}
For hyperparameters of the models, we set the maximum output token length to 10 to prevent redundant responses, while keeping other hyperparameters at their default values.
We employed in-context learning with prompts and did not perform any fine-tuning on the models.

\subsection{Tasks}
\label{ssec:task}

We compare two tasks in our experiments, Abduction task and Deduction task.
Table~\ref{tab:prompts} shows example prompts of each task.

Abduction task provides sentences for \textsf{Rule} and \textsf{Observation} and asks to choose the most plausible hypothesis.
Given a hypothesis $H$, The answer is selected from three options: 
$H$, the negation of $H$, and ``Neither is a good explanation,'' as shown in Table~\ref{tab:prompts}.
In a similar way, the Deduction task provides two premise sentences and three options for the conclusion.

We conducted experiments on the Abduction and Deduction tasks in both zero-shot and few-shot settings \cite{brown2020language}.
For the few-shot prompts, we included eight examples using the same set of terms, corresponding to the eight patterns of abduction shown in Table~\ref{tab:abduction-scheme}, which were inserted between the task description and the problem.
Details of the few-shot prompts can be found in Table~\ref{tab:few-shot-examples-abduction} and Table~\ref{tab:few-shot-examples-deduction} in the Appendix.

We have also tested alternative prompts, which are shown in Table~\ref{tab:alt-prompts} in the Appendix.
However, since there was no performance improvement compared to the prompts listed in Table~\ref{tab:prompts}, they were not adopted.

\begin{table}
\centering

\caption{Example prompts for the Abduction task and Deduction task.}

\scalebox{0.63}{
\begin{minipage}[t]{28em}
\begin{quoteboxa}{Input (Abduction Task)}
\tt
Based on Rule and Observation, from a logical perspective, select the most reasonable hypothesis that explains why Observation holds true.\\
Choose one from the following options (1-3) and answer with the corresponding number.\\
Note that there is a logical relationship between the Rule, Observation, and Hypothesis, where the Observation is logically derived from the Rule and Hypothesis.
\\
[\normalbaselineskip]
Rule: All things that were in the bag are white.\\
Observation: These balls are white.\\
[\normalbaselineskip]
Hypothesis:\\
1. These balls were in the bag.\\
2. These balls were not in the bag.\\
3. Neither is a good explanation.\\
[\normalbaselineskip]
The answer is:

\end{quoteboxa}
\end{minipage}
}
\scalebox{0.63}{
\begin{minipage}[t]{28em}
\begin{quoteboxa}{Input (Deduction Task)}
\tt
Select a sentence that serves as a conclusion based on the following two premises.\\
Choose one from the following options (1-3) and answer with the corresponding number.
\\
[\normalbaselineskip]
P1: All things that were in the bag are white.\\
P2: These balls are white.\\
[\normalbaselineskip]
1. These balls were in the bag.\\
2. These balls were not in the bag.\\
3. Neither.\\
[\normalbaselineskip]
The answer is:
\end{quoteboxa}
\end{minipage}
}

\label{tab:prompts}
\end{table}

\begin{table}
\caption{Accuracy (\%) on the Abduction task ($n = 216$).}

\resizebox{\textwidth}{!}{
\begin{tabular}{llccccccc} \toprule
\textbf{Condition} & \textbf{Model} & \textbf{Overall} & \textbf{Positive} & \textbf{Negative} & \textbf{Neither} & \textbf{Consistent} & \textbf{Inconsistent} & \textbf{Neutral} \\ \midrule
\multirow{4}{*}{Zero-Shot}
& GPT-3.5 & 31.02 & 48.15 & 100.00 &	0.93 & 31.82 & 27.27 & 33.33 \\
& GPT-4 & 41.67 & 80.25 & 92.59 & 0.00 & 46.97 & 34.85 & 42.86 \\
& Llama3-8B & 37.50 & 61.73 & 66.67 & 12.04 & 42.42 & 27.27 & 41.67 \\ 
& Llama3-70B & 37.04 & 64.20 & 100.00 & 0.93 & 45.45 & 31.82 & 34.52 \\ \midrule
\multirow{4}{*}{Few-Shot}
& GPT-3.5 & 29.63 & 44.44 & 96.30 & 1.85 & 33.33 & 22.73 & 32.14 \\
& GPT-4 & 28.70 & 65.43 & 22.22 & 2.78 & 31.82 & 19.70 & 33.33 \\ 
& Llama-3-8B & 28.70 & 41.98 & 100.00 & 0.93 & 33.33 & 27.27 & 26.19 \\ 
& Llama-3-70B & 75.46 & 90.12 & 81.48 & 62.96 & 74.24 & 72.73 & 78.57 \\ \bottomrule 
\end{tabular}
}
\label{tab:res-abd}
\end{table}

\begin{table}
\caption{Accuracy (\%) on the Deduction task ($n = 216$).
}

\resizebox{\textwidth}{!}{
\begin{tabular}{ll c ccc ccc} \toprule
\textbf{Condition} & \textbf{Model} & \textbf{Overall} & \textbf{Positive} & \textbf{Negative} & \textbf{Neither} & \textbf{Consistent} & \textbf{Inconsistent} & \textbf{Neutral} \\ \midrule
\multirow{4}{*}{Zero-Shot}
& GPT-3.5 & 33.80 &	100.00 & 54.32 & 1.85 & 39.39 & 28.79 & 33.33  \\
& GPT-4 & 72.22 & 100.00 & 100.00 & 44.44 & 74.24 & 68.18 & 73.81  \\ 
& Llama-3-8B & 43.52 & 100.00 & 40.74 & 31.48 & 37.88 & 37.88 & 52.38  \\ 
& Llama-3-70B & 53.24 & 100.00 & 80.25 & 21.30 & 60.61 & 40.91 & 57.14  \\ \midrule 
\multirow{4}{*}{Few-Shot}
& GPT-3.5 & 46.30 & 85.19 & 91.36 & 2.78 & 48.48 & 42.42 & 47.62 \\
& GPT-4 & 95.83 & 100.00 & 96.30 & 94.44 & 100.00 & 92.42 & 95.24 \\ 
& Llama-3-8B & 49.54 & 100.00 & 98.77 & 0.00 & 50.00 & 50.00 & 48.81  \\ 
& Llama-3-70B & 84.72 & 92.59 & 80.25 & 86.11 & 90.91 & 72.73 & 89.29  \\ \bottomrule 

\end{tabular}
}
\label{tab:res-deduc}
\end{table}

\subsection{Results}
Tables~\ref{tab:res-abd} and \ref{tab:res-deduc} show the Abduction and Deduction task results.
The columns labeled \textit{Positive}, \textit{Negative}, and \textit{Neither} correspond to instances where the correct answer is the hypothesis $H$ (positive form), the negation of $H$, and ``Neither is a good explanation,'' respectively.

For the Abduction task in the zero-shot setting, the overall accuracy of the highest-performing model (GPT-4) was around 42\%, which was slightly above the chance level.
While the model achieved over 80\% accuracy on problems with correct answers labeled as \textit{Positive} or \textit{Negative}, it performed poorly on problems where the correct answer was \textit{Neither}.
With regard to the content types, the accuracy of \textit{Inconsistent} problems was around 10\% lower than the other two types (\textit{Consistent} and \textit{Inconsistent}).
This suggests that belief biases are also reproduced in abduction tasks.

In the few-shot setting, Llama-3-70B was the only model that showed a significant performance improvement, achieving approximately 63\% accuracy on the problems whose answer type was \textit{Neither} and around 75\% overall accuracy.
The overall accuracy for GPT-3.5 showed a slight improvement over the zero-shot setting, while for GPT-4, the overall accuracy was lower than in the zero-shot setting. 
The score for problems where the correct answer was \textit{Neither} slightly increased for both GPT models.

For the Deduction task, the few-shot setting improved performance across all models compared to the zero-shot setting, with gains ranging from 6.02 to 31.48 points in the overall accuracy.
GPT-4 was the best-performing model in both settings, achieving an overall accuracy of 72.22\% in the zero-shot setting and 95.83\% in the few-shot setting. However, except for Llama-3-8B, accuracy remained lower for the problems labeled \textit{Inconsistent} compared to those labeled \textit{Consistent} and \textit{Neutral}.

\subsection{Discussion}
\label{ssec:discussion}
\paragraph{\bf Are the models abductive reasoners?}
The results on deduction tasks generally show the similar tendencies to the previous findings~\cite{dasgupta2023language,ando-etal-2023-evaluating,eisape2024systematic}.
That is, LLMs' performance was quite low in the problems whose correct answer were \textit{Neither} and the sentences in the problems contradict common sense belief.
The exception is Llama-3-70B in the few-shot setting;
still the accuracy for abduction (75.46\%) was lower than that for deduction (84.72\%).

It was anticipated that the results for the abduction task would be better than those for the deduction task, as abduction is more akin to everyday human reasoning, whereas deduction requires more reflective reasoning.
However, the results surprisingly showed that the accuracy on abduction tasks was low overall.
In particular, for the problems where the correct answer was \textit{Neither}, LLMs often failed to solve them at all.

Given that abduction has been studied less than deduction, there is a possibility that while deduction cases are included more in the LLMs' training data, abduction problems are fewer.
Also, considering that it is expected that sentences that perform hypothesis selection and hypothesis generation are included in natural texts, it is possible that there is difficulty in applying abduction to the syllogistic form.
The observation human-like belief biases in abduction is consistent with Pereira et al.~\cite{pereira2014contextual}, which reports that belief bias is observed in abductions.

\paragraph{\bf Why do the models tend to mistakenly choose \textit{Negative}?}

In the Abduction task, for problems where \textit{Neither} was the correct answer (108 problems), the distribution of GPT-4's predictions was as follows: 32 problems were answered as \textit{Positive}, 76 as \textit{Negative}, and 0 as \textit{Neither}.
In contrast, in the Deduction task, the distribution was as follows: 26 problems were answered as \textit{Positive}, 34 as \textit{Negative}, and 48 as \textit{Neither}.
Thus, in the Abduction task, there was a more pronounced tendency to incorrectly answer \textit{Neither} problems and choose \textit{Negative} as the correct answer.

To analyze this tendency in more detail, we calculated the rate at which a \textit{Negative} was selected as the hypothesis when ``No'' or ``not'' appears in the \textsf{Rule} or \textsf{Observation}.
In the case of GPT-4,
this rate was 67.90\% for the Abduction task
(with the actual rate of the correct answer being \textit{Negative} at 16.67\%), while it was 70.99\% for the Deduction task (with the actual rate of the correct answer being \textit{Negative} at 50\%).
Thus, in both tasks, \textit{Negative} sentences were selected at a higher rate than the actual rate of correct answers being \textit{Negative}, with this tendency being more pronounced in the Abduction task.
This tendency may be due to an effect similar to the atmosphere effects~\cite{chater1999probability}, where the presence of negation in the \textsf{Rule} or \textsf{Observation} leads to the selection of a hypothesis that also contains negation.

\paragraph{\bf Do the models answer the problems  as deduction?}

We examined the scores of abduction problems when labeled as if they were deduction problems. 
For example, an affirmation of the antecedent (e.g., \textit{All B are C, A is B}) has no correct hypothesis for abduction, and therefore the correct answer is \textit{Neither}. However, if it is conceived as deduction, it logically entails \textit{A is C} and is labeled as \textit{Positive}.
When comparing the correct labels for deduction to GPT-4's predictions, the agreement rate was 51.85\% for \textit{Overall}, 100.0\% for \textit{Positive}, 93.83\% for \textit{Negative}, and 8.33\% for \textit{Neither}.
In \textit{Overall}, this was about 10 points higher than the accuracy for the original abduction labels.
For example, among inferences of the form ``affirmation of the antecedent,'' all inferences of the form \textit{All B are C, A is B} led to the selection of the correct deductive answer, and 89\% of inferences of the form \textit{No B are C, A is B} also led to the selection of the correct deductive answer. 

This suggests that LLMs are influenced by deduction when solving abduction problems.
However, in general, the agreement rate does not reach the level of accuracy in the Deduction task (falling short by about 20 points for \textit{Overall} and by about 35 points for \textit{Neither}), suggesting that LLMs are not completely mistaking abduction problems for deduction problems.

\paragraph{\bf Does the word ``Hypothesis'' mislead by suggesting entailment relationship?}
In Natural Language Inference (NLI) tasks~\cite{williams-etal-2018-broad}, the conclusion that entails the premise is usually called ``Hypothesis.''
Given this fact, we investigated the possibility that the word ``Hypothesis'' itself does not function as a hypothesis to explain \textsf{Observation}, but instead it suggests an entailment or deductive relation.
In particular, substituting the term ``Hypothesis'' with ``Reason'' had little effect on improving the score.

\paragraph{\bf Do the models choose contradictory answers?}
To specify error tendencies, we investigated whether LLMs choose the answer regardless of logical consistency. 
We checked whether the LLMs choose the answer that contradicts the given \textsf{Rule} or \textsf{Observation}, but few such cases are observed.

\section{Conclusion and Future Work}
In this paper, we created a dataset to test abductive reasoning abilities of LLMs and compared LLM's accuracy on abductive reasoning tasks with deductive reasoning tasks.
The results showed that LLMs performed worse in abduction than in deduction. In addition, human-like belief biases are observed in abduction as well as in deduction.

Abduction is considered to be a more ordinary inference than deduction is.
Therefore, it is expected that humans would perform better for abduction than for deduction.
This expected tendency is different from the tendency of LLMs, as shown in this paper.
Comparisons between LLMs and humans for the abductive reasoning tasks, as well as further investigations on the error tendencies or biases in abductive reasoning through these comparisons, are topics for future work.

We adopted Peirce's initial characterization of abduction in syllogistic framework and focused on hypothesis generation tasks.
Although three options, \textit{Positive}, \textit{Negative}, and \textit{Neither}, are included in each problem, the options other than the correct answer do not serve as hypothesis from the logical perspective, so the task can be seen as a simpler version of hypothesis generation, rather than a task of choosing the best one from multiple hypotheses.
However, other characterizations are also possible.
Abduction can be understood as \textit{Inference to the Best Explanation}. 
Tasks such as selecting the best hypothesis that explains the premises from the plural candidates that are already logical explanations are expected in future work.
Also, although we characterized abduction with syllogistic framework, it can be understood as a probable reasoning.
Comparison of humans and LLMs by a probabilistic (Bayesian) approach to abduction is an area for future work.
Furthermore, evaluating more complex types of reasoning, such as extended syllogisms and conditionals, are also left for future work.

\subsection*{Acknowledgements}

We thank the anonymous reviewers for their helpful comments and suggestions, which have improved the paper.
This work is partially supported by JST, CREST Grant Number JPMJCR2114,
JST BOOST, Japan Grant Number JPMJBS2409,
the KGRI Challenge Grant from the Keio University Global Research Institute, and JSPS Kakenhi Grant Numbers JP24K00004, JP21K00016, JP21H00467, JP23K20416, and JP21K18339.

\bibliographystyle{splncs04}
\bibliography{baroco}

@inproceedings{
ozeki2024exploring,
title={Exploring Reasoning Biases in Large Language Models Through Syllogism: Insights from the {N}eu{BAROCO} Dataset},
author={Ozeki, Kentaro and Ando, Risako  and
      Morishita, Takanobu  and
      Abe, Hirohiko  and
      Mineshima, Koji  and
      Okada, Mitsuhiro},
booktitle={Findings of the Association for Computational Linguistics: ACL 2024},
year={2024}
}

@book{pohl2012cognitive,
  title={Cognitive Illusions: A Handbook on Fallacies and Biases in Thinking, Judgement and Memory},
  author={Pohl, R{\"u}diger F},
  year={2022},
  edition={3},
  publisher={Routledge},
  doi={https://doi.org/10.4324/9780203720615}
}

@article{llama3modelcard,
title={{Llama 3 Model Card}},
author={AI@Meta},
year={2024}
}

@article{wang2019superglue,
  title={Superglue: A stickier benchmark for general-purpose language understanding systems},
  author={Wang, Alex and Pruksachatkun, Yada and Nangia, Nikita and Singh, Amanpreet and Michael, Julian and Hill, Felix and Levy, Omer and Bowman, Samuel},
  journal={Advances in neural information processing systems},
  volume={32},
  year={2019}
}

@article{kojima2022large,
  title={Large language models are zero-shot reasoners},
  author={Kojima, Takeshi and Gu, Shixiang Shane and Reid, Machel and Matsuo, Yutaka and Iwasawa, Yusuke},
  journal={Advances in neural information processing systems},
  volume={35},
  pages={22199--22213},
  year={2022}
}

@article{eisape2024systematic,
      title={A Systematic Comparison of Syllogistic Reasoning in Humans and Language Models}, 
      author={Tiwalayo Eisape and MH Tessler and Ishita Dasgupta and Fei Sha and Sjoerd van Steenkiste and Tal Linzen},
      doi={10.48550/arXiv.2311.00445},
      year={2024},
      journal={arXiv preprint arXiv:2311.00445}
}

@inproceedings{bhagavatula2020abductive,
title={Abductive Commonsense Reasoning},
author={Chandra Bhagavatula and Ronan Le Bras and Chaitanya Malaviya and Keisuke Sakaguchi and Ari Holtzman and Hannah Rashkin and Doug Downey and Wen-tau Yih and Yejin Choi},
booktitle={International Conference on Learning Representations},
year={2020}
}

@book{peirceCollectedPapers,
  title     = {Collected Papers of Charles Sanders Peirce},
  editor    = {Charles Hartshorne and Paul Weiss and Arthur W. Burks},
  publisher = {Harvard University Press},
  address   = {Cambridge, Massachusetts},
  year      = {1931--1958},
  volumes   = {8},
  note      = {Volumes 1--6 edited by Charles Hartshorne and Paul Weiss, 1931--1935; volumes 7--8 edited by Arthur W. Burks, 1958}
}

@InCollection{douven-sep-abduction,
	author       =	{Douven, Igor},
	title        =	{{Abduction}},
	booktitle    =	{The {Stanford} Encyclopedia of Philosophy},
	editor       =	{Edward N. Zalta},
	howpublished =	{\url{https://plato.stanford.edu/archives/sum2021/entries/abduction/}},
	year         =	{2021},
	edition      =	{{S}ummer 2021},
	publisher    =	{Metaphysics Research Lab, Stanford University}
}

@book{Douven2022-DOUTAO-5,
	author = {Igor Douven},
	publisher = {The MIT Press},
	title = {The Art of Abduction},
	year = {2022},
    doi = {10.7551/mitpress/14179.001.0001}
}

@article{wei2022chain,
  title={Chain-of-thought prompting elicits reasoning in large language models},
  author={Wei, Jason and Wang, Xuezhi and Schuurmans, Dale and Bosma, Maarten and Xia, Fei and Chi, Ed and Le, Quoc V and Zhou, Denny and others},
  journal={Advances in Neural Information Processing Systems},
  volume={35},
  pages={24824--24837},
  year={2022}
}

@article{ouyang2022gpt3,
  title={Training language models to follow instructions with human feedback},
  author={Ouyang, Long and Wu, Jeffrey and Jiang, Xu and Almeida, Diogo and Wainwright, Carroll and Mishkin, Pamela and Zhang, Chong and Agarwal, Sandhini and Slama, Katarina and Ray, Alex and others},
  journal={Advances in Neural Information Processing Systems},
  volume={35},
  pages={27730--27744},
  year={2022}
}

@article{openai2023gpt4,
  title={{GPT}-4 Technical Report},
  author={OpenAI},
  journal={arXiv preprint arXiv:2303.08774},
  doi={10.48550/arXiv.2303.08774},
  year={2023}
}

@article{dasgupta2023language,
      title={Language models show human-like content effects on reasoning tasks}, 
      author={Ishita Dasgupta and Andrew K. Lampinen and Stephanie C. Y. Chan and Hannah R. Sheahan and Antonia Creswell and Dharshan Kumaran and James L. McClelland and Felix Hill},
      journal={arXiv preprint arXiv:2207.07051},
      doi={10.48550/arXiv.2207.07051},
      year={2023}
}

@article{aghahadi2022avicenna,
  title={Avicenna: a challenge dataset for natural language generation toward commonsense syllogistic reasoning},
  author={Aghahadi, Zeinab and Talebpour, Alireza},
  journal={Journal of Applied Non-Classical Logics},
  volume={32},
  number={1},
  pages={55--71},
  year={2022},
  doi={10.1080/11663081.2022.2041352},
  publisher={Taylor \& Francis}
}

@inproceedings{gubelmann2022philosophically,
  title={A Philosophically-Informed Contribution to the Generalization Problem of Neural Natural Language Inference: Shallow Heuristics, Bias, and the Varieties of Inference},
  author={Gubelmann, Reto and Niklaus, Christina and Handschuh, Siegfried},
  booktitle={Proceedings of the 3rd Natural Logic Meets Machine Learning Workshop (NALOMA III)},
  pages={38--50},
  year={2022}
}

@article{dong2020learning,
  title={Learning Syllogism with {Euler} Neural-Networks},
  author={Dong, Tiansi and Li, Chengjiang and Bauckhage, Christian and Li, Juanzi and Wrobel, Stefan and Cremers, Armin B},
  journal={arXiv preprint arXiv:2007.07320},
  doi={10.48550/arXiv.2007.07320},
  year={2020}
}

@book{manktelow1999reasoningEng,
  title={Reasoning and Thinking},
  year={1999},
  author={Manktelow, Ken},
  publisher={Psychology Press}
}

@inproceedings{ando-etal-2023-evaluating,
    title = "Evaluating Large Language Models with {N}eu{BAROCO}: Syllogistic Reasoning Ability and Human-like Biases",
    author = "Ando, Risako  and
      Morishita, Takanobu  and
      Abe, Hirohiko  and
      Mineshima, Koji  and
      Okada, Mitsuhiro",
    booktitle = "Proceedings of the 4th Natural Logic Meets Machine Learning Workshop",
    year = "2023",
    pages = "1--11"
}

@inproceedings{devlin-etal-2019-bert,
    title = "{BERT}: Pre-training of Deep Bidirectional Transformers for Language Understanding",
    author = "Devlin, Jacob  and
      Chang, Ming-Wei  and
      Lee, Kenton  and
      Toutanova, Kristina",
    booktitle = "Proceedings of the 2019 Conference of the North {A}merican Chapter of the Association for Computational Linguistics",
    year = "2019",
    pages = "4171--4186"
}

@article{brown2020language,
  title={Language models are few-shot learners},
  author={Brown, Tom and Mann, Benjamin and Ryder, Nick and Subbiah, Melanie and Kaplan, Jared D and Dhariwal, Prafulla and Neelakantan, Arvind and Shyam, Pranav and Sastry, Girish and Askell, Amanda and others},
  journal={Advances in neural information processing systems},
  volume={33},
  pages={1877--1901},
  year={2020}
}

@inproceedings{williams-etal-2018-broad,
    title = "A Broad-Coverage Challenge Corpus for Sentence Understanding through Inference",
    author = "Williams, Adina  and
      Nangia, Nikita  and
      Bowman, Samuel",
    booktitle = "Proceedings of the 2018 Conference of the North {A}merican Chapter of the Association for Computational Linguistics: Human Language Technologies, Volume 1 (Long Papers)",
    year = "2018",
    doi = "10.18653/v1/N18-1101",
    pages = "1112--1122"
}

@article{chater1999probability,
  title={The probability heuristics model of syllogistic reasoning},
  author={Chater, Nick and Oaksford, Mike},
  journal={Cognitive Psychology},
  volume={38},
  number={2},
  pages={191--258},
  doi={10.1006/cogp.1998.0696},
  year={1999}
}

@book{evans1989bias,
  title={Bias in Human Reasoning: Causes and Consequences.},
  author={Evans, Jonathan St.B. T.},
  year={1989},
  publisher={Lawrence Erlbaum Associates, Inc}
}

@article{Geurts2003-GEURWQ,
	year = {2003},
	journal = {Cognition},
	volume = {86},
	number = {3},
	author = {Bart Geurts},
	pages = {223--251},
	title = {Reasoning with Quantifiers},
        doi = {10.1016/S0010-0277(02)00180-4}
}

@incollection{FriedmanForthcoming-FRIZE,
	author = {Jane Friedman},
	booktitle = {Towards an Expansive Epistemology: Norms, Action, and the Social Sphere},
	editor = {Baron Reed and A. K. Flowerree},
	publisher = {Routledge},
	title = {Zetetic Epistemology},
	year = {forthcoming}
}

@incollection{Hookway2006-HOOEAI,
	author = {Christopher Hookway},
	booktitle = {Epistemology Futures},
	editor = {Stephen Hetherington},
	pages = {95--110},
	publisher = {Oxford University Press},
	title = {Epistemology and Inquiry: The Primacy of Practice},
        doi = {10.1093/oso/9780199273317.003.0006},
	year = {2006}
}

@article{Hookway2008-HOOQEA,
	author = {Christopher Hookway},
	journal = {Grazer Philosophische Studien},
	number = {1},
	pages = {1--21},
	publisher = {Editions Rodopi},
	title = {Questions, Epistemology, and Inquiries},
	volume = {77},
        doi = {10.1163/18756735-90000841},
	year = {2008}
}

@article{pereira2014contextual,
  title={Contextual abductive reasoning with side-effects},
  author={Pereira, Luis Moniz and Dietz, Emmanuelle-Anna and H{\"o}lldobler, Steffen},
  journal={Theory and practice of logic programming},
  volume={14},
  number={4-5},
  pages={633--648},
  year={2014},
  doi={10.1017/S1471068414000258},
  publisher={Cambridge University Press}
}

@article{Medianovskyi2022-MEDOEA-2,
	author = {Kyrylo Medianovskyi and Ahti{-}Veikko Pietarinen},
	journal = {Philosophies},
	number = {2},
	pages = {35},
	title = {On Explainable {AI} and Abductive Inference},
	volume = {7},
        doi={10.3390/philosophies7020035},
	year = {2022}
}

@Inbook{Bellucci2022,
author="Bellucci, Francesco
and Pietarinen, Ahti-Veikko",
editor="Magnani, Lorenzo",
title="Peirce's Abduction",
bookTitle="Handbook of Abductive Cognition",
doi="10.1007/978-3-031-10135-9_7",
year="2022",
publisher="Springer International Publishing",
address="Cham",
pages="1--14",
isbn="978-3-030-68436-5",
}

@inproceedings{Stilgenbauer2017ReasoningSF,
  title={Reasoning Strategies for Diagnostic Probability Estimates in Causal Contexts: Preference for Defeasible Deduction over Abduction},
  author={Jean-Louis Stilgenbauer and Jean Baratgin and Igor Douven},
  booktitle={Proceedings of the 4th International Workshop on Defeasible and Ampliative Reasoning (DARe-17)},
  year={2017}
}

@article{Stilgenbauer2019AssessingTA,
  title={Assessing the accuracy of diagnostic probability estimation: Evidence for defeasible modus ponens},
  author={Jean-Louis Stilgenbauer and Jean Baratgin},
  journal={International Journal of Approximate Reasoning},
  year={2019},
  volume={105},
  doi = {10.1016/j.ijar.2018.11.015},
  pages={229-240}
}

\appendix

\section{Details of Prompts}

Table~\ref{tab:alt-prompts} shows examples of prompts we tested that scored lower than the prompts we finally adopted.
Tables~\ref{tab:few-shot-examples-abduction} and \ref{tab:few-shot-examples-deduction} show the prompts with eight exemplars (8-shot prompts) used in the few-shot setting.

\begin{table}[h!]
\centering

\caption{Examples of alternative prompts not adopted.}

\scalebox{0.63}{
\begin{minipage}[t]{28em}
\begin{quoteboxa}{Input (Abduction Task)}
\tt
Suppose the following Observation is logically derived from Rule and Hypothesis.\\
Choose the most appropriate sentence for the Hypothesis from the following options (1-3) and answer with the corresponding number.\\
[\normalbaselineskip]
Rule: All things that were in the bag are white.\\
Hypothesis: ???\\
----------------------------------------\\
Observation: These balls are white.\\
[\normalbaselineskip]
1. These balls were in the bag.\\
2. These balls were not in the bag.\\
3. Neither.\\
[\normalbaselineskip]
The answer is:
\end{quoteboxa}
\end{minipage}
\hspace{2em}
\begin{minipage}[t]{28em}
\begin{quoteboxa}{Input (Abduction Task)}
\tt
You are an inquirer. You know that the following Rule holds true in the world. Additionally, you have recently confirmed that the following Observation is also true. \\
Given this information, you want to discover the mechanism behind why these hold true.\\

Based on the Rule and Observation below, select the most plausible hypothesis from a logical perspective that explains why the Observation is valid. \\
Please respond with the corresponding number from the numbers 1-3.
\\
[\normalbaselineskip]
Rule: All things that were in the bag are white.\\
Observation: These balls are white.\\
[\normalbaselineskip]
Hypothesis:\\
1. These balls were in the bag.\\
2. These balls were not in the bag.\\
3. Neither is a good explanation.\\
[\normalbaselineskip]
The answer is:
\end{quoteboxa}
\end{minipage}
}
\label{tab:alt-prompts}
\end{table}

\begin{table}[h!]
\centering

\caption{An example few-shot prompt for the Abduction task.}

\scalebox{0.9}{
\begin{quoteboxa}{Input (Abduction Task)}
\begin{multicols}{2}
\scriptsize
\tt
Based on Rule and Observation, from a logical perspective, select the most reasonable hypothesis that\\ explains why Observation holds true.\\
Choose one from the following options (1-3) and answer with the corresponding number.\\
Note that there is a logical relationship between the Rule, Observation, and Hypothesis, where the\\ Observation is logically derived from the Rule and Hypothesis.\\
[\normalbaselineskip]
Rule: All things that are sold at the shop are waterproof.\\
Observation: These shoes are waterproof.\\
[\normalbaselineskip]
Hypothesis:\\
1. These shoes are sold at the shop.\\
2. These shoes are not sold at the shop.\\
3. Neither is a good explanation.\\
[\normalbaselineskip]
The answer is: 1\\
[\normalbaselineskip]
Rule: All things that are waterproof are sold at the shop.\\
Observation: These shoes are waterproof.\\
[\normalbaselineskip]
Hypothesis:\\
1. These shoes are sold at the shop.\\
2. These shoes are not sold at the shop.\\
3. Neither is a good explanation.\\
[\normalbaselineskip]
The answer is: 3\\
[\normalbaselineskip]
Rule: All things that are sold at the shop are waterproof.\\
Observation: These shoes are not waterproof.\\
[\normalbaselineskip]
Hypothesis:\\
1. These shoes are sold at the shop.\\
2. These shoes are not sold at the shop.\\
3. Neither is a good explanation.\\
[\normalbaselineskip]
The answer is: 3\\
[\normalbaselineskip]
Rule: All things that are waterproof are sold at the shop.\\
Observation: These shoes are not waterproof.\\
[\normalbaselineskip]
Hypothesis:\\
1. These shoes are sold at the shop.\\
2. These shoes are not sold at the shop.\\
3. Neither is a good explanation.\\
[\normalbaselineskip]
The answer is: 2\\
[\normalbaselineskip]
Rule: No things that are sold at the shop are waterproof.\\
Observation: These shoes are waterproof.\\
[\normalbaselineskip]
Hypothesis:\\
1. These shoes are sold at the shop.\\
2. These shoes are not sold at the shop.\\
3. Neither is a good explanation.\\
[\normalbaselineskip]
The answer is: 3\\
[\normalbaselineskip]
Rule: No things that are waterproof are sold at the shop.\\
Observation: These shoes are waterproof.\\
[\normalbaselineskip]
Hypothesis:\\
1. These shoes are sold at the shop.\\
2. These shoes are not sold at the shop.\\
3. Neither is a good explanation.\\
[\normalbaselineskip]
The answer is: 3\\
[\normalbaselineskip]
Rule: No things that are sold at the shop are waterproof.\\
Observation: These shoes are not waterproof.\\
[\normalbaselineskip]
Hypothesis:\\
1. These shoes are sold at the shop.\\
2. These shoes are not sold at the shop.\\
3. Neither is a good explanation.\\
[\normalbaselineskip]
The answer is: 1\\
[\normalbaselineskip]
Rule: No things that are waterproof are sold at the shop.\\
Observation: These shoes are not waterproof.\\
[\normalbaselineskip]
Hypothesis:\\
1. These shoes are sold at the shop.\\
2. These shoes are not sold at the shop.\\
3. Neither is a good explanation.\\
[\normalbaselineskip]
The answer is: 1\\
[\normalbaselineskip]
Rule: All things that were in the bag are white.\\
Observation: These balls are white.\\
[\normalbaselineskip]
Hypothesis:\\
1. These balls were in the bag.\\
2. These balls were not in the bag.\\
3. Neither is a good explanation.\\
[\normalbaselineskip]
The answer is:
\end{multicols}
\end{quoteboxa}
} 

\label{tab:few-shot-examples-abduction}

\end{table}

\begin{table}[h!]
\centering

\caption{An example few-shot prompt for the Deduction task.}

\scalebox{0.9}{
\begin{quoteboxa}{Input (Deduction Task)}
\begin{multicols}{2}
\scriptsize
\tt
Select a sentence that serves as a conclusion based on the following two premises.\\
Choose one from the following options (1-3) and answer with the corresponding number.\\
[\normalbaselineskip]
P1: All things that are sold at the shop are waterproof.\\
P2: These shoes are waterproof.\\
[\normalbaselineskip]
1. These shoes are sold at the shop.\\
2. These shoes are not sold at the shop.\\
3. Neither.\\
[\normalbaselineskip]
The answer is: 3\\
[\normalbaselineskip]
P1: All things that are waterproof are sold at the shop.\\
P2: These shoes are waterproof.\\
[\normalbaselineskip]
1. These shoes are sold at the shop.\\
2. These shoes are not sold at the shop.\\
3. Neither.\\
[\normalbaselineskip]
The answer is: 1\\
[\normalbaselineskip]
P1: All things that are sold at the shop are waterproof.\\
P2: These shoes are not waterproof.\\
[\normalbaselineskip]
1. These shoes are sold at the shop.\\
2. These shoes are not sold at the shop.\\
3. Neither.\\
[\normalbaselineskip]
The answer is: 2\\
[\normalbaselineskip]
P1: All things that are waterproof are sold at the shop.\\
P2: These shoes are not waterproof.\\
[\normalbaselineskip]
1. These shoes are sold at the shop.\\
2. These shoes are not sold at the shop.\\
3. Neither.\\
[\normalbaselineskip]
The answer is: 3\\
[\normalbaselineskip]
P1: No things that are sold at the shop are waterproof.\\
P2: These shoes are waterproof.\\
[\normalbaselineskip]
1. These shoes are sold at the shop.\\
2. These shoes are not sold at the shop.\\
3. Neither.\\
[\normalbaselineskip]
The answer is: 2\\
[\normalbaselineskip]
P1: No things that are waterproof are sold at the shop.\\
P2: These shoes are waterproof.\\
[\normalbaselineskip]
1. These shoes are sold at the shop.\\
2. These shoes are not sold at the shop.\\
3. Neither.\\
[\normalbaselineskip]
The answer is: 2\\
[\normalbaselineskip]
P1: No things that are sold at the shop are waterproof.\\
P2: These shoes are not waterproof.\\
[\normalbaselineskip]
1. These shoes are sold at the shop.\\
2. These shoes are not sold at the shop.\\
3. Neither.\\
[\normalbaselineskip]
The answer is: 3\\
[\normalbaselineskip]
P1: No things that are waterproof are sold at the shop.\\
P2: These shoes are not waterproof.\\
[\normalbaselineskip]
1. These shoes are sold at the shop.\\
2. These shoes are not sold at the shop.\\
3. Neither.\\
[\normalbaselineskip]
The answer is: 3\\
[\normalbaselineskip]
P1: All things that were in the bag are white.\\
P2: These balls are white.\\
[\normalbaselineskip]
1. These balls were in the bag.\\
2. These balls were not in the bag.\\
3. Neither.\\
[\normalbaselineskip]
The answer is:

\end{multicols}
\end{quoteboxa}
} 

\label{tab:few-shot-examples-deduction}

\end{table}

\end{document}